\documentclass[10pt,twocolumn,letterpaper]{article}

\usepackage{cvpr}      
\usepackage{colortbl}
\usepackage{xcolor}
\usepackage{multirow}
\usepackage{makecell}
\usepackage{adjustbox}
\usepackage{algorithm}
\usepackage{algorithmicx}
\usepackage{algpseudocode}
\usepackage{amsmath}
\usepackage{footmisc}
\usepackage{tabularx}
\floatname{algorithm}{Algorithm} 

\definecolor{lightred}{RGB}{251, 205, 209}
\definecolor{tgreen}{RGB}{202, 207, 224}

\definecolor{cvprblue}{rgb}{0.21,0.49,0.74}
\usepackage[pagebackref,breaklinks,colorlinks,allcolors=cvprblue]{hyperref}

\def\paperID{4129} 
\def\confName{CVPR}
\def\confYear{2025}

\title{DCEvo: Discriminative Cross-Dimensional Evolutionary Learning for Infrared and Visible Image Fusion }

\author{
  Jinyuan Liu\textsuperscript{\rm 1}, \quad Bowei Zhang\textsuperscript{\rm 2}, \quad Qingyun Mei\textsuperscript{\rm 2}, \quad Xingyuan Li\textsuperscript{\rm 2}, \quad Yang Zou\textsuperscript{\rm 3},\\  Zhiying Jiang\textsuperscript{\rm 4}, \quad Long Ma\textsuperscript{\rm 2}, \quad Risheng Liu\textsuperscript{\rm 2}, \quad Xin Fan\textsuperscript{\rm 2}\thanks{Corresponding author.}\\
  {\small\textsuperscript{1} School of Mechanical Engineering, Dalian University of Technology }\\
  {\small\textsuperscript{2} School of Software Technology \& DUT-RU International School of ISE, Dalian University of Technology }\\
  {\small\textsuperscript{3} School of Computer Science, Northwestern Polytechnical University } \\
  {\small\textsuperscript{4} College of Information Science and Technology, Dalian Maritime University }\\
  {\tt\small atlantis918@hotmail.com} \hspace{0.1cm}
  {\tt\small xin.fan@dlut.edu.cn} \hspace{0.1cm}
}

\let\oldtwocolumn\twocolumn
\renewcommand\twocolumn[1][]{%
    \oldtwocolumn[{#1}{
    \begin{center}
           \includegraphics[width=0.99\textwidth,trim={10pt 5pt 0pt 35pt}]{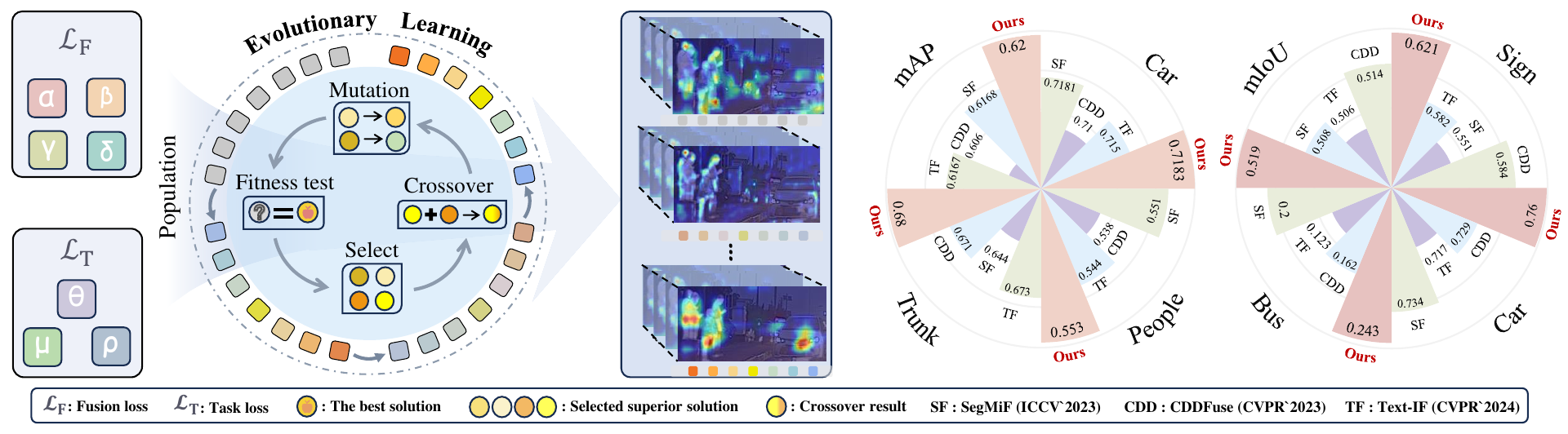}
           \captionof{figure}{The left part presents an illustration of our evolutionary learning strategy. It can cooperatively adjust the loss function, optimizing the model with both fusion performance and its downstream task precision with the evolution of loss function. 
           The right part shows the accuracy comparison of downstream tasks with state-of-the-art methods. Our approach performs the highest accuracy in multiple metrics.  }
           \label{fig:impact}
        \end{center}
    }]
}
\begin{document}
\maketitle
\begin{abstract}
Infrared and visible image fusion integrates information from distinct spectral bands to enhance image quality by leveraging the strengths and mitigating the limitations of each modality. Existing approaches typically treat image fusion and subsequent high-level tasks as separate processes, resulting in fused images that offer only marginal gains in task performance and fail to provide constructive feedback for optimizing the fusion process. To overcome these limitations, we propose a Discriminative Cross-Dimension Evolutionary Learning Framework, termed DCEvo, which simultaneously enhances visual quality and perception accuracy. Leveraging the robust search capabilities of Evolutionary Learning, our approach formulates the optimization of dual tasks as a multi-objective problem by employing an Evolutionary Algorithm (EA) to dynamically balance loss function parameters. Inspired by visual neuroscience, we integrate a Discriminative Enhancer (DE) within both the encoder and decoder, enabling the effective learning of complementary features from different modalities. Additionally, our Cross-Dimensional  Embedding (CDE) block facilitates mutual enhancement between high-dimensional task features and low-dimensional fusion features, ensuring a cohesive and efficient feature integration process. Experimental results on three benchmarks demonstrate that our method significantly outperforms state-of-the-art approaches, achieving an average improvement of  9.32\%  in visual quality while also enhancing subsequent high-level tasks. The code is available at \url{https://github.com/Beate-Suy-Zhang/DCEvo}.
\end{abstract}

\section{Introduction}
\label{sec:intro}

Infrared images capture thermal radiation, functioning well in darkness, and through smoke or fog, but suffer from low resolution and limited textures \cite{El_Ahmar_2022_CVPR, li2024contourlet}. Conversely, visible  images offer high resolution and rich colors with detailed edges, yet falter in low light or adverse weather. Consequently, fusing infrared and visible images enhances overall image quality and significantly improves information acquisition in complex environments such as autonomous driving~\cite{SegMIF_ICCV2023}, remote sensing~\cite{Yu_2023_CVPR}, security surveillance~\cite{TextIF_CVPR2024}, medical imaging~\cite{CDDFuse_CVPR2023}, and military reconnaissance \cite{TarDAL_CVPR2022}.

To meet diverse application needs, Infrared and Visible Image Fusion (IVIF) aims to create high-quality visual images through pixel-level enhancements, thereby improving clarity and detail for human observers \cite{U2_2022, TextIF_CVPR2024}. Additionally, this fusion technique enhances the accuracy of downstream perception tasks, such as object detection and scene analysis, providing task-level improvements \cite{Geng_2024_CVPR, liu2024promptfusion}. By achieving both pixel and task-level enhancements, IVIF supports intelligent systems in better environment understanding and informed decision making. 

In recent years, rapid advancements in deep learning have significantly propelled the field of IVIF. Deep learning–based methods have markedly surpassed traditional approaches in fusion performance \cite{Zheng_2024_CVPR}. 
Despite these advancements, the majority of IVIF approaches still focus on improving visual aesthetics rather than enhancing the effectiveness of subsequent high-level vision tasks, which are vital for numerous practical applications \cite{SegMIF_ICCV2023}. For instance, existing strategies generally handle image fusion and object detection as distinct problems, treating detection solely as an afterthought. Consequently, the resulting fused images provide only marginal gains in detection accuracy, and the detection outcomes fail to offer insights that could further optimize the fusion process \cite{MetaFusion_CVPR2023, TarDAL_CVPR2022, Alpher46}.

Instead of adopting a divide-and-conquer approach to handle tasks independently, addressing this integrated problem introduces several challenges. \textbf{Challenges in Multi-Dimensional Feature Coordination}: Each objective leverages distinct information during model inference, meaning that features beneficial for image fusion may not be suitable for the subsequent high-level tasks, and vice versa. \textbf{Difficulties in Loss Function Adjustment}: A common strategy to achieve dual objectives simultaneously involves linking networks with a combined weighted loss function. The success of both goals heavily depends on precise parameter tuning. However, this manual method often enhances one objective at the expense of the other, preventing the attainment of optimal results for both tasks.

To address the  \textbf{Multi-Dimensional Feature Coordination} and \textbf{Loss Function Adjustment} challenges in IVIF, this paper proposes a Discriminative Cross-Dimensional Evolutionary Learning Framework for harnessing the visual enhancement and accurate perception. 
Inspired by the powerful search capabilities of Evolutionary Learning, we introduce a novel approach to address the challenges of collaborative optimization between dual task learning processes. We frame this as a multi-objective optimization problem, utilizing an Evolutionary Algorithm (EA) to update the coefficients of cross-task constraints and learn the optimal parameters for each task.
To support this framework, we first design a Discriminative Enhancer (DE) based on principles from visual neuroscience and its inherent physical properties. The DE is seamlessly integrated into both the Encoder and Decoder components of our model. Following this, we establish a Cross-Dimensional Embedding (CDE) block that embeds object-aware features generated by high-dimensional tasks into our low-dimensional fusion features.
Our contributions are detailed in four  aspects as follows:
\begin{itemize}

\item We propose a Discriminative Cross-Dimensional Evolutionary Learning (DCEvo) approach for the fusion of infrared and visible images, resulting in visually appealing fused images that maintain high accuracy in task performance. To the best of our knowledge, this is the first study to integrate evolutionary learning into the field of IVIF.

\item To effectively learn complementary features and enhance the distinctive characteristics of different modalities, we introduce a discriminative enhancer that is seamlessly integrated into both the encoder and decoder.

\item We establish a cross-dimensional embedding block, where high-dimensional task features can supervise low-dimensional fusion features, and vice versa. This design facilitates comprehensive interaction between features of different dimensions, thereby enabling mutual enhancement of both fusion and downstream tasks.

\item We propose an evolutionary learning strategy to progressively identify hyperparameters balancing two tasks within the loss function. This approach overcomes the challenges associated with empirical hyperparameter tuning by systematically optimizing the parameters to achieve equilibrium between the competing objectives.

\end{itemize}

\begin{figure*}[t]
  \centering
\includegraphics[width=0.99\linewidth,trim={10pt 5pt 10pt 20pt}]{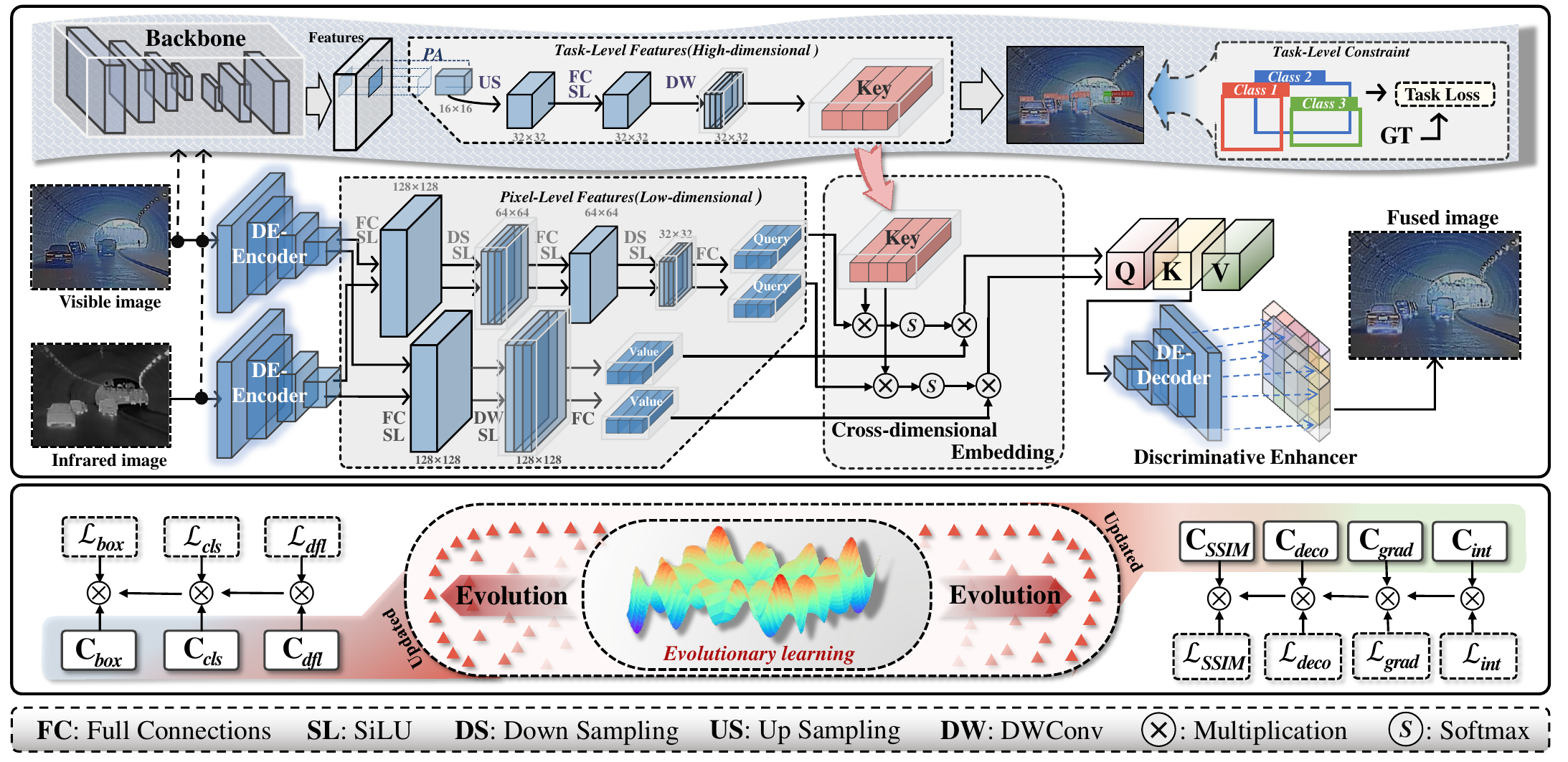}

   \caption{An Overall Illustration of our DCEvo architecture. 
The middle depicts our infrared and visible image fusion network to generate images by coupling pixel-level feature and task-level feature.
The upper part denotes the detection network, which embeds the task-level feature   for fusion supervision to enable that fused images contain object information. In the cooperative training process of detection and fusion network, we propose an evolutionary learning strategy to search the coefficient of the optimization objectives, as  in the bottom part.
}
   \label{fig:pipeline}
\end{figure*}

\section{Related work}
\label{sec:formatting}
\subsection{Learning-based IVIF approaches}
IVIF has driven the development of advanced methodologies for multi-modal feature processing, leveraging rapid advancements in deep learning. Techniques such as Autoencoders (AE) \cite{Alpher36,Alpher37, Alpher38}, Convolutional Neural Networks (CNN) \cite{Alpher31, Alpher32,  9349250, Alpher34, Alpher35}, Generative Adversarial Networks (GAN) \cite{Alpher39, Alpher40}, Transformers \cite{liu2024coconet, Ma_SwinFusion, CDDFuse_CVPR2023}, and Diffusion Models \cite{DDFM_2023_ICCV} have significantly enhanced IVIF performance. For instance, CDDFuse~\cite{CDDFuse_CVPR2023} combines a decomposition network with a Transformer-CNN architecture to extract both global and local features for effective image fusion. Similarly, U2Fusion~\cite{U2_2022} utilizes a dense network with data-driven weights to evaluate the importance of input images during fusion. 

In recent years, there has been a growing emphasis on task-oriented IVIF methods \cite{TarDAL_CVPR2022, MetaFusion_CVPR2023, SegMIF_ICCV2023}. TarDAL~\cite{TarDAL_CVPR2022} integrates an object detection network within a cascade framework, optimizing both networks simultaneously for detection-oriented outcomes. MetaFusion~\cite{MetaFusion_CVPR2023} introduces a detection feature-guided fusion network that learns semantic information to enhance fusion quality. Additionally, SegMiF~\cite{SegMIF_ICCV2023} incorporates a segmentation sub-network to improve pixel-level features for both fusion and subsequent segmentation tasks. More methods are surveyed in \cite{10812907}.

\subsection{Evolutionary learning}

Evolutionary learning, rooted in biological evolution theory, mimics natural evolutionary processes to address complex optimization challenges. Utilizing mechanisms such as selection, crossover, and mutation, it iteratively refines candidate solutions toward optimality. This approach includes heuristic algorithms like Genetic Algorithms (GA) \cite{Alpher06}, Ant Colony Optimization (ACO) \cite{Alpher07}, and Particle Swarm Optimization (PSO) \cite{Alpher08}, which are extensively applied in deep learning. Due to its superior performance in optimization tasks, evolutionary learning is widely employed in diverse fields, including disease detection \cite{Alpher09,Alpher10,Alpher11}, path planning \cite{Alpher12,Alpher13}, and image fusion \cite{Alpher14,Alpher15}.

Multi-objective optimization problems (MOPs) involve simultaneously optimizing multiple but often conflicting objectives, resulting in a set of Pareto optimal solutions that represent various trade-offs. Evolutionary learning excels in solving MOPs robustly through its global search capabilities and population-based evolution without gradient flow \cite{Alpher16,Alpher17}. This effectiveness also makes it a popular choice for hyperparameter optimization \cite{Alpher19,Alpher20,Alpher21}.

\section{Motivation}
Evolutionary learning offers significant advantages in complex multi-task optimization, including neural architecture search \cite{Alpher19}, hyperparameter tuning \cite{Alpher21}, and multi-objective optimization \cite{Alpher16}. In collaborative optimization of image fusion and downstream perception, severe challenges arise due to typically discrete problems where existing methods fall into local optima, limiting overall performance. Evolutionary learning excels by simulating natural selection and genetic mechanisms to explore vast solution spaces and identify global optima, converting discrete problems into continuous optimization. We utilize evolutionary learning to adaptively adjust hyperparameters based on the loss functions of image fusion and target detection, employing multi-objective optimization to balance their needs, surpassing individual task optimization.

\renewcommand{\thealgorithm}{1} 
    \begin{algorithm}
        \caption{EA for hyperparameter optimization
    \label{alg:GA}} 
        \begin{algorithmic}[1] 
            \Require Number of population $N$, Number of iterations $I_n$, Probability of population individual mutating $P_m$
            \Ensure Global best solution $G_b$
            \State Start with a random initial population $P_1$
            \State \text{Calculate initial population {fitness} $F_1$} 
            
            \For{t = 1 to $I_n$}
                
                \State \text{\textbf{Sort} (population, descending, $F_t$)}
                
                \State \text{\textbf{Select} individuals with high fitness for crossover }
                \State NewIndividuals $\longleftarrow$ \textbf{Crossover}()
                \Statex \quad \quad $Q_t \longleftarrow $ NewIndividuals

                \For{k = 1 to N}
                    \State $x \sim U(0,100)$
                    \State \textbf{if} x $< P_m$ \textbf{then}
                        \State \quad \quad Mutation(k-th individual)
                    \State \textbf{end if}
                \EndFor

                \State NewIndividuals $\longleftarrow$ \textbf{Mutate}()
                \State $Q_t \longleftarrow Q_t \cup $ NewIndividuals
                
                \State \text{Calculate population $M_t$ fitness}
                
                \State \text{Calculate \textbf{selected} probability of each individual  }
                \Statex \quad \quad $p\left(x_i\right)=\frac{\text { fitness }\left(x_i\right)}{\sum_{j=1}^{\left|M_t\right|} \operatorname{fitness}\left(x_j\right)}, \quad \forall x_i \in M_t$
                \State \text{Selecting a new generation of population through}
                \Statex \text{\quad \quad  the \textbf{roulette wheel}, $r \sim U(0,1)$}
                 
                \Statex \quad \quad $P_{t+1}=\left \{x_i\mid \sum_{i=1}^{k-1} p(x_i)\le r\le \sum_{i=1}^{k}p(x_i ) \right\}$ 
    
            \EndFor            
            \State \Return $G_b$
        \end{algorithmic}
    \end{algorithm}

\section{Methodology}
\label{sec:method}

This section details our DCEvo framework, including the learning of evolutionary hyperparameters, the discriminative enhancer, and cross-dimensional feature embeddings, as illustrated in Figure~\ref{fig:pipeline}.

\begin{figure*}[t]
  \centering
\includegraphics[width=\linewidth, height=120pt, trim={10pt 0pt 10pt 0pt}]{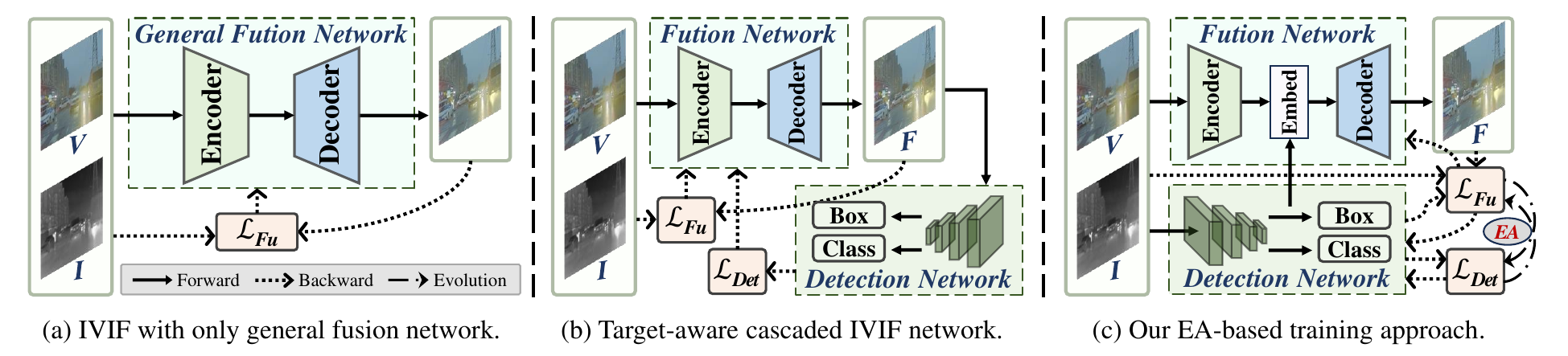}

   \caption{The illustration of different network workflows and learning strategies of IVIF towards upper-level tasks. Strategy (a) trains  fusion network by only low-level constraints, while (b) cascades a detection network to guide the fusion network with additional high-level constraints. Our (c) training approach deploys an evolutionary algorithm to optimize the two task cooperatively and effectively.}
   \label{fig:difference}
\end{figure*}

\subsection{Learning evolutionary hyperparameters}
Selecting model hyperparameters remains a significant challenge, traditionally relying on prior knowledge. However, separately optimizing two networks based on distinct priors fails to simultaneously satisfy the requirements of both fusion and downstream tasks. Additionally, the total loss function comprises multiple components, creating a multi-objective problem (MOP) where conflicts may arise between different elements. Fortunately, evolutionary learning can concurrently explore diverse regions of the solution space and identify multiple non-dominated solutions in a single run, making it well-suited for MOPs. Therefore, we propose an \textbf{Evolutionary Algorithm (EA)} to optimize loss function coefficients for improved training, as illustrated in Figure~\ref{fig:difference}(c). 
Our EA employs a genetic algorithm that evaluates each solution’s fitness based on its loss value—the lower the loss, the higher the fitness. High-fitness individuals are selected for crossover, increasing the proportion of superior solutions within the population. Mutation introduces random variations to explore a broader solution space, and each iteration selects high-fitness individuals via roulette wheel selection for the next generation.
The procedure is presented in Algorithm \ref{alg:GA}.

We give the notation of the model formulation. Infrared, visible, and fused images are denoted as $\mathbf{I}_{ir},\mathbf{I}_{vis},\mathbf{I}_{f}$. In the optimization  of image fusion, to obtain fused images with high visual quality, multiple constraints are applied. The loss function is 
$ \mathcal{L}_{Fu}= \mathcal{L}_{SSIM}+ \mathcal{L}_{deco}+ \mathcal{L}_{grad}+ \mathcal{L}_{int}$, 
where the 
$ \mathcal{L}_{SSIM}$ $=$ $ \mathcal{L}_{SSIM}(\mathbf{I}_{ir}, {\mathbf{I}}_{f}) + \mathcal{L}_{SSIM}(\mathbf{I}_{vis}, {\mathbf{I}}_{f}) $, $ \mathcal{L}_{int}=\frac{1}{HW}\left \| \mathbf{I}_f-max(\mathbf{I}_{ir},\mathbf{I}_{vis}) \right \|  $  and $ \mathcal{L}_{grad}=\frac{1}{HW}\left \| \left | \nabla \mathbf{I}_f \right | -max(\left | \nabla \mathbf{I}_{ir} \right | ,\left | \nabla \mathbf{I}_{vis} \right | ) \right \|  $. The $ \mathcal{L}_{deco}$ is proposed in \cite{CDDFuse_CVPR2023} to correlate basic features from  cross-modal inputs.

The loss function of the object detection part consists of three parts, i.e., $\mathcal{L}_{Det} = \mathcal{L}_{cls} + \mathcal{L}_{box} + \mathcal{L}_{dfl}$. The first part is the classification loss, which is calculated as follows:
\begin{equation}
     \mathcal{L}_{cls}=-\sum_{i=1}^{N_{cls}}[\mathbf{y}_i\log(1+e^{-\hat{\mathbf{y}}_i})+(1-\mathbf{y}_i)\log (1+e^{\hat{\mathbf{y}}_i} )],
\end{equation}
where the $\hat{\mathbf{y}}$ is the predict class label, and the $\mathbf{y}$ is the true label. The $N_{cls}$ is the number of classes.
The second part is the Complete Intersection Over Union (CIOU) loss, which is calculated as follows:
\begin{equation}
\mathcal{L}_{box} = 1 - (IOU - \frac{d_2^2}{d_C^2} - \frac{v^2}{(1 - IOU) + v}),
\end{equation}
where~$v = \frac{4}{\pi^2}(\text{arctan}\frac{w_g}{h_g} - \text{arctan}\frac{w_p}{h_p})^2, $ $d_2$ is the Euclidean distance between the center points of the two boxes, and $d_C$ is the diagonal distance of the smallest bounding rectangle box. The $w_g$ and  $h_g$ are the width and height of the ground truth box, while $w_p$ and  $h_p$ are the width and height of the predicted box.
The third part is the distribution focal loss $\mathcal{L}_{dfl}$, which is used to rapidly focus  on the label box \cite{DFL_2023}. 

\subsection{Discriminative enhancer}
In IVIF, the feature extraction network is essential, yet the inherent properties of feature maps are rarely addressed. To this end, we introduce a \textbf{Discriminative Enhancer (DE)} to enhance the feature representation, thereby boosting the learning efficiency of networks.

In the feature representation, every feature map  contains a special semantic information for  objects. 
To formulate the object and background features, for a single feature map 
$\mathbf{{X}} \in \mathbb{R}^{{H} \times {W}}$,  assume  $\mathbf{{X}}$ $=$ ${\mathbf{X}}_{{o}}$ $+$ $\mathbf{{X}}_{{b}}$,
where 
$\mathbf{{X}}_{{o}}$
and $\mathbf{{X}}_{{b}}$ 
are the feature maps that only contain object pixels and background pixels. 
The pixel values in $\mathbf{{X}},$ $
\mathbf{{X}}_{{o}}, $
$\mathbf{{X}}_{{b}}$ are in [0, 1].
The mean value of  $\mathbf{{X}}$ is  
$\mu$ = $\frac{1}{M}\sum_{i=1}^{M}{x_i}$, where $x_i$ denotes a pixel in the feature map $\mathbf{{X}}$ and   $M={H}$ $\times$ ${W}$. 
Therefore, it is obvious to enlarge the value of difference 
$||\mathbf{{X}}_{{o}} - \mathbf{{X}}_{{b}}||_1$.  
The mean value of  $\mathbf{{X}}_{{o}}$ is 
$\mu_o$ = $\frac{1}{M_o}\sum_{i=1}^{M_o}{x_{o,i}}$ $\in$ ($\mu$, 1], 
while the  mean value of  $\mathbf{{X}}_{{b}}$ is 
$\mu_b$ = $\frac{1}{M_b}\sum_{i=1}^{M_b}{x_{b,i}}$ $\in$ [0, $\mu$).
Since the objects with temperature are highlighted in IVIF, 
we assume that $\mu_o$ $>$ $\mu$ $>$ $\mu_b$.
Therefore, to modulate the surroundings for objects, we can enlarge the difference
$D(\mathbf{{X}})$ $=$ $\sum_{i=1}^{M}|x_i - \mu|$.
Due to  $\mu_o > \mu > \mu_b$, it can be easily seen that  
$D(\mathbf{{X}})$  $<$ $\sum_{i=1}^{M}|\frac{x_i^2}{\mu} - \mu|$, which indicate that the value of $\frac{x_i^2}{\mu}$ can serve more modulation than $x_i$.

Note that this formulation needs the condition $\mathbf{{X}}$, 
$\mathbf{{X}}_{{o}}$, 
$\mathbf{{X}}_{{b}} > 0$.
the Sigmoid activation $S(x)$ is used for importance value mapping because of its monotonicity and boundedness. 
Therefore, we use the Sigmoid activation to map the pixel values to $y_i$ $=$ $S(x_i)$ $\in$ $(0,1)$ whose mean value is $\mu_y$. 
We give the importance weight for pixel $x_i$ for feature enhancement.
Every enhanced  pixel in the enhanced feature map $\widetilde{\mathbf{{X}}}$ according to the surrounding of $\mathbf{{X}}$  is as:
\begin{equation}
\widetilde{x_i} = x_i \times S(\frac{y_i^2}{\mu_y}),
\end{equation}
where $S(\cdot)$ denotes the Sigmoid activation.

we use DE in both fusion encoder and  decoder, which is as the  surround modulation for objects to generate fused images with prominent-object information. 

\subsection{Cross-dimensional feature embedding}
Apart from the prominent object representation, we propose a  \textbf{Cross-Dimensional  Embedding (CDE)} method, supervising the fusion network to integrate the cross-task feature.

Encoder-decoder networks effectively represent multi-modal features. However, they overlook the impact of feature representations across different tasks. Task-level models can extract high-dimensional information. For example, in object detection, feature pyramid networks distinguish objects from their surroundings, providing semantic information for the detection head. Therefore, our DCEvo method leverages detection features by using the CDE module to integrate features from both tasks.

 As depicted in Figure~\ref{fig:pipeline}, the fusion encoder converts the infrared image $\mathbf{I}_{ir}$ and the visible image $\mathbf{I}_{vis}$ to features with $\mathbf{F}_{ir}$ and $\mathbf{F}_{vis}$ respectively, while the detection network transforms the $\mathbf{I}_{vis}$ into $\mathbf{F}_{det}$.  Different from feature embedding in other tasks, image feature fusion requires the feature maps of two inputs contain the same scene entity. To properly match the represented feature, we conduct Patch Align (PA) by slicing the part of input fusion patch from the detection feature map $\mathbf{F}_{det}$. The sliced patch of $\mathbf{F}_{det}$ is $\mathbf{F}_{pdet}$. Then, $\mathbf{F}_{pdet}$,  $\mathbf{F}_{ir}$ and $\mathbf{F}_{vis}$ are fed into  separate CNN blocks to obtain $\widehat{\mathbf{F}}_{pdet}$,  $\widehat{\mathbf{F}}_{ir}$ and $\widehat{\mathbf{F}}_{vis}$ for feature embedding. Let $\{\mathbf{Q}$, $\mathbf{K}$, $\mathbf{V}\}$ $=$ $\{\widehat{\mathbf{F}}_{pdet}$, $\widehat{\mathbf{F}}_{ir}$, $\widehat{\mathbf{F}}_{vis}\}$, and then  our cross-dimensional embedding operation output $\widehat{\mathbf{F}}_{ca}$ is:
\begin{equation}
\widehat{\mathbf{F}}_{ca} = \mathcal{C}\mathcal{D}\mathcal{E}(\mathbf{Q}, \mathbf{K}, \mathbf{V}),
\end{equation}
where $\mathcal{C}\mathcal{D}\mathcal{E}(\cdot)$ denotes the cross-dimensional embedding.

Through the above feature embedding, the detection features are integrated into image fusion network, enabling object awareness for fused images. Subsequently, the feature maps $\widehat{\mathbf{F}}_{ca}$ are fed into the self-attention module. Finally, the fused images are generated by the image decoder.

\begin{figure}[t]
  \centering
\includegraphics[width=\linewidth, height=140pt,trim={0pt 0pt 0pt 0pt}]{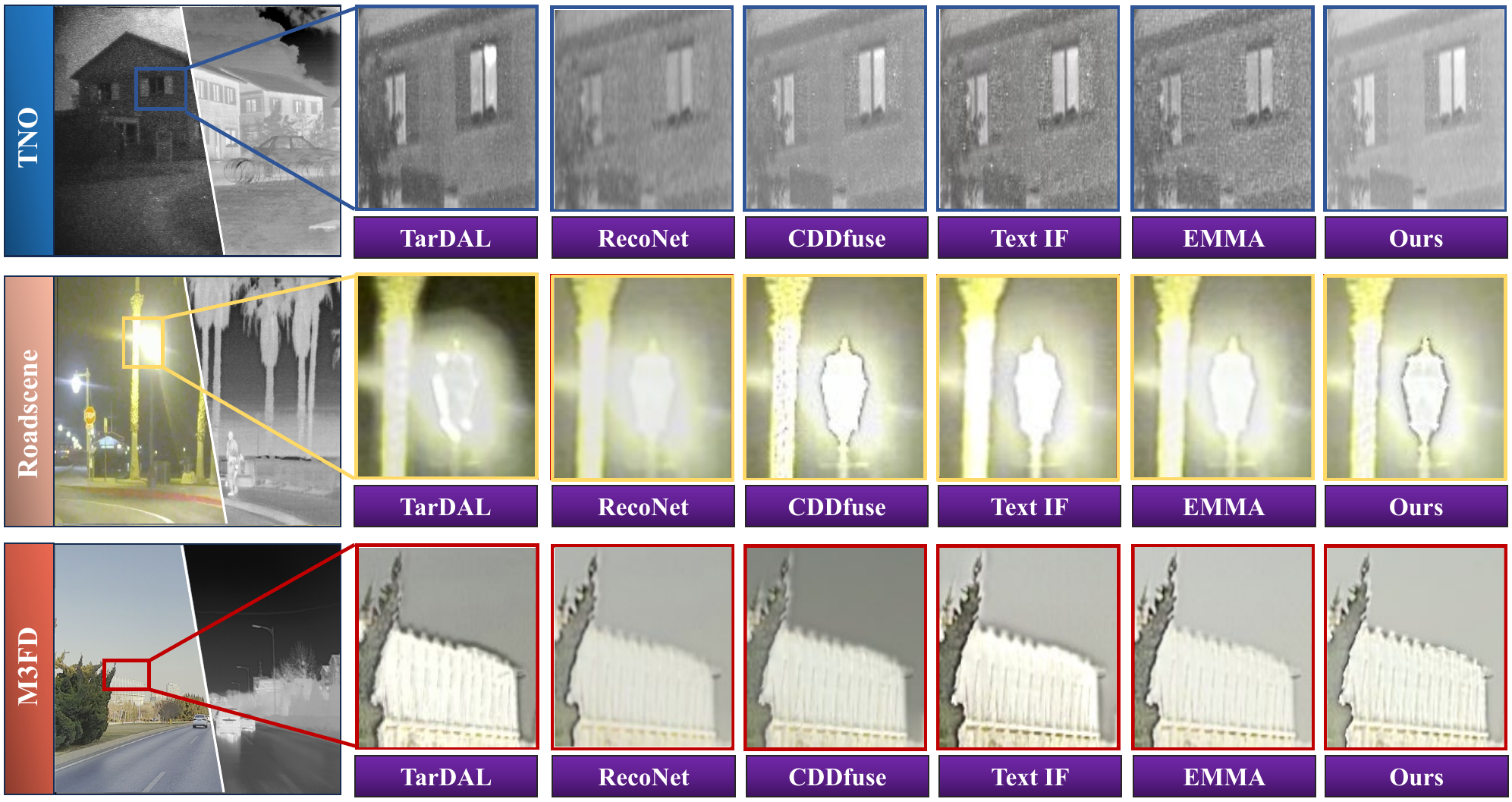}

   \caption{Qualitative comparisons  of our DCEvo and existing image fusion methods. From top to bottom: low-light in TNO, high-brightness in RoadScene and low-quality in M$^3$FD.
}
   \label{fig:fusionresult}
\end{figure}


\begin{table*}[h!]
\centering
\footnotesize
    \begin{adjustbox}{max width=\textwidth}
\begin{tabular}{c|cccc|cccc|cccc|cccc}
\Xhline{1.1pt}

\multicolumn{1}{c|}{Datasets} & \multicolumn{4}{c|}{M\(^{3}\)FD} & \multicolumn{4}{c|}{RoadScene} & \multicolumn{4}{c|}{TNO} & \multicolumn{4}{c}{FMB} \\ 
\cmidrule(lr){1-1} \cmidrule(lr){2-5} \cmidrule(lr){6-9} \cmidrule(lr){10-13} \cmidrule(lr){14-17}
\multicolumn{1}{c|}{Methods} & MI & SSIM & VIF & Qabf & MI & SSIM & VIF & Qabf & MI & SSIM & VIF & Qabf & MI & SSIM & VIF & Qabf \\  
\hline

DDFM 	& 2.91 & 0.46 & 0.70 & 0.48 
& \cellcolor{tgreen!100}1.84 & \cellcolor{tgreen!100}0.04 & \cellcolor{tgreen!100}0.21 & \cellcolor{tgreen!100}0.14 
& \cellcolor{tgreen!50}1.94 & \cellcolor{tgreen!100}0.22 & \cellcolor{tgreen!100}0.40 & \cellcolor{tgreen!100}0.25 
& 3.10 & 0.30 & 0.68 & 0.44 \\ 

YDTR 	& 3.18 & 0.47 & 0.70 & 0.48 
& 3.02 & 0.48 & 0.69 & 0.44 
& 2.82 & 0.50 & 0.71 & 0.41 
& 3.21 & 0.30 & 0.68 & \cellcolor{tgreen!50}0.43 \\ 

ReCoNet 	& 3.06 & 0.43 & \cellcolor{tgreen!50}0.62 & 0.49 
& 3.09 & 0.45 & 0.64 & 0.38 
& 2.57 & 0.42 & \cellcolor{tgreen!50}0.57 & 0.35 
& 3.17 & 0.33 & 0.66 & 0.50 \\ 

TarDAL 	& 3.16 & 0.44 & 0.63 & \cellcolor{tgreen!100}0.41 
& 3.34 & 0.45 & 0.62 & 0.43 
& 2.85 & 0.49 & 0.65 & 0.40 
& 3.31 & 0.31 & \cellcolor{tgreen!50}0.60 & \cellcolor{tgreen!100}0.36 \\ 

LRRNet 	& 2.80 & \cellcolor{tgreen!50}0.39 & 0.63 & 0.50 
& 2.79 & \cellcolor{tgreen!50}0.33 & \cellcolor{tgreen!50}0.57 & \cellcolor{tgreen!50}0.35 
& 2.66 & 0.43 & 0.62 & 0.36 
& 2.90 & \cellcolor{tgreen!50}0.28 & 0.65 & 0.48 \\ 

MetaFusion& \cellcolor{tgreen!100}2.36 & \cellcolor{tgreen!100}0.38 & \cellcolor{red!20}\underline{0.87} & \cellcolor{tgreen!100}0.41 
& \cellcolor{tgreen!50}2.28 & 0.41 & 0.59 & 0.39 
& \cellcolor{tgreen!100}1.71 & \cellcolor{tgreen!50}0.38 & 0.70 & \cellcolor{tgreen!50}0.29 
& \cellcolor{tgreen!100}2.48 & 0.33 & 0.77 & \cellcolor{tgreen!50}0.43 \\ 

SwinFusion& \cellcolor{red!20}\underline{4.16} & 0.50 & 0.84 & 0.61 
& 3.34 & \cellcolor{red!20}\underline{0.49} & \cellcolor{red!20}\underline{0.73} & 0.46 
& 3.36 & \cellcolor{red!20}\underline{0.52} & 0.81 & 0.51 
& 3.09 & 0.33 & 0.78 & 0.59 \\ 

TIM 		& \cellcolor{tgreen!50}2.74 & 0.40 & \cellcolor{tgreen!100}0.61 & 0.50 
& \cellcolor{red!20}\underline{3.62} & 0.44 & 0.70 & 0.39 
& \cellcolor{red!20}\underline{3.79} & 0.47 & 0.78 & 0.44
 & \cellcolor{tgreen!50}2.60 & \cellcolor{tgreen!100}0.27 & \cellcolor{tgreen!100}0.58 & 0.49 \\  

CDDFuse 	& 3.90 & \cellcolor{red!45}\textbf{0.51} & 0.85 & 0.61 
& 3.08 & 0.48 & 0.71 & 0.48 
& 3.23 & \cellcolor{red!20}\underline{0.52} & 0.84 & {0.52} 
& 3.63 & \cellcolor{red!20}\underline{0.35} & 0.80 & 0.58 \\ 

SegMiF 	& 3.05 & 0.48 & 0.85 & \cellcolor{red!20}\underline{0.65} 
& 2.75 & 0.47 & 0.71 & 0.53 
& 3.04 & 0.49 & \cellcolor{red!20}\underline{0.87} & \cellcolor{red!45}\textbf{0.57} 
& \cellcolor{red!20}\underline{3.73} & 0.34 & 0.79 & 0.57 \\  

EMMA 	& 3.81 & 0.45 & 0.82 & 0.59 
& 3.18 & 0.46 & 0.70 & 0.45 
& 2.98 & 0.48 & 0.75 & 0.46 
& 3.69 & \cellcolor{red!20}\underline{0.35} & 0.80 & 0.58 \\  

Text-IF 	& 3.68 & 0.47 & \cellcolor{red!45}\textbf{0.94} & \cellcolor{red!45}\textbf{0.68} 
& 2.88 & 0.46 & \cellcolor{red!20}\underline{0.73} & \cellcolor{red!20}\underline{0.54} 
& 3.14 & 0.48 & 0.82 & 0.52 
& 3.37 & 0.34 & \cellcolor{red!20}\underline{0.83} & \cellcolor{red!20}\underline{0.61} \\

\hline

 {DCEvo} & 
\cellcolor{red!45} \textbf{4.54} & \cellcolor{red!45}\textbf{0.51} & \cellcolor{red!20}\underline{0.87} & {0.64} & 
\cellcolor{red!45}\textbf{3.69} & \cellcolor{red!45}\textbf{0.51} & \cellcolor{red!45}\textbf{0.80} & \cellcolor{red!45}\textbf{0.57} & 
\cellcolor{red!45}\textbf{4.22} & \cellcolor{red!45}\textbf{0.54} & \cellcolor{red!45}\textbf{0.91} & \cellcolor{red!45}\textbf{0.57} &
\cellcolor{red!45}\textbf{4.38} & \cellcolor{red!45}\textbf{0.49} & \cellcolor{red!45}\textbf{0.97} & \cellcolor{red!45}\textbf{0.68}\\ 
\Xhline{1.1pt}
\end{tabular}
    \end{adjustbox}
\caption{Quantitative comparison of infrared and visible image fusion between our DCEvo and state-of-the-art methods on \(\text{M}^\text3\text{FD}\), {RoadScene}, TNO and FMB datasets. \textbf{Boldface} shows the best while \underline{underline} shows the second-best results. \textcolor{red!45}{Red}, \textcolor{red!20}{light red}, \textcolor{tgreen!100}{blue}, and \textcolor{tgreen!50}{light blue} represent the best, second-best, worst, and second-worst results, respectively.}
\label{tb1:ivif}
\end{table*}

\cellcolor{red!45}
\cellcolor{red!20}
\cellcolor{tgreen!50}
\cellcolor{tgreen!100}

\begin{figure*}[t]
  \centering
\includegraphics[width=\linewidth, height=145pt ,trim={0pt 0pt 0pt 0pt}]{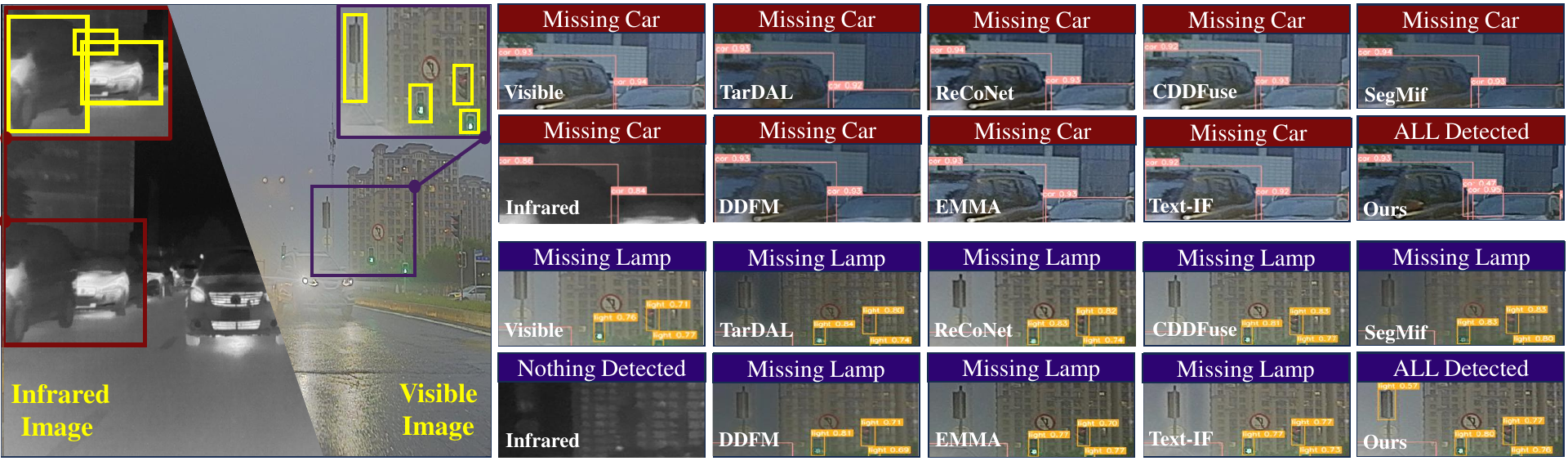}

   \caption{Qualitative comparison task of our method and existing infrared and visible image fusion methods in downstream object detection on the M$^3$FD dataset. The objects in our fusion images are fully detected.
}
   \label{fig:detvis}
\end{figure*}

\section{Experiments}

To evaluate the performance of our proposed methods, we conduct experiments on four datasets, where four (M\(^{3}\)FD~\cite{TarDAL_CVPR2022}, RoadScene~\cite{RoadSceneDATA}, TNO~\cite{TNODATA} and FMB~\cite{SegMIF_ICCV2023}) for IVIF, one (M\(^{3}\)FD) for object detection and one (FMB) for image segmentation. 
Our designed image fusion model of DCEvo are firstly pretrained on MSRS datasets.
Then, the  DCEvo  is trained on the M$^3$FD datasets by evolutionary learning where model trainer are updated every epoch. In our evolutionary algorithm, the population size, number of iterations, and mutation rate are set to 5, 5, and 10, respectively. 
These images are fed to the object detection model and semantic segmentation model for training.   
All experiments are conducted with the PyTorch.

\subsection{Infrared and visible image fusion}

We evaluate the IVIF performance of DCEvo without CDE by comparing with 12 state-of-the-art methods, including YDTR~\cite{YDTR_2023}, SwinFusion~\cite{Ma_SwinFusion}, ReCoNet~\cite{ReCoNet_ECCV2022}, TarDAL~\cite{TarDAL_CVPR2022},
SegMiF~\cite{SegMIF_ICCV2023}, DDFM~\cite{DDFM_2023_ICCV}, LRRNet~\cite{LRRNet_TPRMI2023}, CDDFuse~\cite{CDDFuse_CVPR2023}, MetaFusion~\cite{MetaFusion_CVPR2023}, TIM~\cite{Alpher46}, EMMA~\cite{EMMA_CVPR2024}, Text-IF~\cite{TextIF_CVPR2024}.

\noindent\textbf{Qualitative Comparisons}. Figure~\ref{fig:fusionresult} presents qualitative results from four datasets. Our method demonstrates significant advantages over existing approaches. Firstly, DCEvo produces fused images with reduced noise levels. Secondly, thermal objects are distinctly highlighted, ensuring clear and accurate representation.

\noindent\textbf{Quantitative Comparisons}. We compare our DCEvo with the methoned 12 counterparts  on M\(^{3}\)FD, RoadScene, TNO and FMB datasets (containing 300, 221, 57 and 280 IVIF images respectively), as shown in Table~\ref{tb1:ivif}. 
On the RoadScene, TNO and FMB datasets, our method shows best on all the four metrics, especially showing significant advantages in MI, SSIM and VIF, showing our method can generate high visual quality images. On the M\(^{3}\)FD dataset, our method is also competitive, showing highest  MI and SSIM. These promising results indicate that our DCEvo maintains and performs higher fidelity, more edge information, less distortion with input images.

\begin{table*}[h!]
    \centering
\footnotesize
    \begin{adjustbox}{max width=\textwidth}
    \begin{tabular}{c|ccccccc|ccccccc}
\Xhline{1.1pt}
\multicolumn{1}{c|}{Datasets} & \multicolumn{7}{c|}{M\(^{3}\)FD} & \multicolumn{7}{c}{FMB} \\
\cmidrule(lr){1-1} \cmidrule(lr){2-8} \cmidrule(lr){9-15} 

Method & People & Car & Bus & Light & Moto & Trunk & mAP & Car & Person & Sky & Bus & Motor& Pole & mIoU \\
\hline





ReCoNet & 54.07 & 71.24 & 74.17 & 53.48 &  44.18 & 66.78 & 60.66 

&  \cellcolor{red!45} \textbf{76.86} & 67.12 & \cellcolor{tgreen!50} 91.43 & 12.91 & 21.55 & 37.96 & \cellcolor{tgreen!100} 48.98 \\

YDTR & 54.82 & 71.37 & 74.02 & 52.82 & 48.22 & 67.52 & 61.46 

& \cellcolor{tgreen!100} 70.08 & \cellcolor{tgreen!100} 64.85 & 92.25 & \cellcolor{tgreen!50} 11.98 & 23.48 & 39.13 & 49.99 \\

SwinFusion & 53.61 & 71.06 & 74.06 & 53.25 & 46.19 &  {68.02} & 61.03 

& 72.38 & 67.91 & 92.68 & 22.12 & 20.15 & 40.24 & 50.70 \\

DDFM & 54.35 & 71.42 & 75.18 & 54.07 & 46.45 & \cellcolor{red!20} \underline{68.06} & 61.59 

& 72.20 & 68.59 & 92.00 & 17.74 & \cellcolor{tgreen!50} 13.64 & \cellcolor{red!20} \underline{41.27} & 50.88 \\

CDDFuse & 53.85 & 71.03 & 74.44 & \cellcolor{tgreen!100} 51.82 & 45.40 & 67.10 & 60.60 

& 72.93 & 66.92 & 93.06 & 16.22 & 23.36 & 38.14 & 51.43 \\

{MetaFusion} & \cellcolor{tgreen!50} 52.80 & \cellcolor{tgreen!50} 70.50 & 74.15 &  52.66 & 45.04 & \cellcolor{tgreen!50} 65.90 & \cellcolor{tgreen!50} 60.18  

& 73.53 & 68.50 & 93.32 &  \cellcolor{red!20}  \underline{26.02} & 23.85 & 37.87 & \cellcolor{red!20} {\underline{51.79}} \\

TIM & \cellcolor{tgreen!100} 51.30 & \cellcolor{tgreen!100} 70.44 & 74.83 & \cellcolor{tgreen!50} 52.29 & \cellcolor{tgreen!50} 44.17 & \cellcolor{tgreen!100} 65.34 & \cellcolor{tgreen!100} 59.69 

& 74.64 & 66.80 & \cellcolor{tgreen!100} 90.56 & 21.48 & 14.78 & 37.69 & 51.31 \\

LRRNet & 54.27 & \cellcolor{red!20} \underline{71.82} & 74.43 & 54.25 & 47.10 & \cellcolor{red!45} \textbf{68.41} & \cellcolor{red!20} \underline{61.71} 

& 71.87 & 68.25 & 92.80 & 19.76 & 15.16 & 39.56 & 50.42 \\

TarDAL & \cellcolor{red!45} \textbf{55.44} &  70.58 & \cellcolor{tgreen!100} 73.17 & 52.92 & \cellcolor{tgreen!100} 42.82 & 66.16 & \cellcolor{tgreen!50} 60.18 

&  \cellcolor{red!20} \underline{76.04} & \cellcolor{tgreen!50} 64.95 & 92.11 & \cellcolor{tgreen!100} 9.63 & 19.88 & \cellcolor{tgreen!100} 34.82 & \cellcolor{tgreen!50} 49.39 \\

SegMiF & \cellcolor{red!20} \underline{55.08} & \cellcolor{red!20} \underline{71.82} & \cellcolor{red!45} \textbf{76.69} & 53.91 & 46.18 & 66.46 & 61.69
 
& 73.40 &  65.98 & 92.93 & 20.00 & \cellcolor{tgreen!100} 11.71 & 38.24 & 50.81 \\

EMMA & 54.66 & 71.70 & \cellcolor{tgreen!50} 73.69 & \cellcolor{red!20} \underline{54.64} & 47.04 & 67.88 & 61.60 

& \cellcolor{tgreen!50} 71.55 &  \cellcolor{red!20} \underline{68.82} & 92.51 & 19.19 &  \cellcolor{red!45} \textbf{27.22} & \cellcolor{tgreen!50} 36.63 & 51.55 \\

TextIF & 54.37 & 71.48 & 75.56 & 52.91 & \cellcolor{red!20} {48.39} & 67.35 & 61.68 

& 71.78 & 67.69 &  \cellcolor{red!45} \textbf{93.57} & 12.35 &  \cellcolor{red!20} \underline{26.49} & 41.07 & 50.60 \\
\hline

DCEvo & {54.02} & \cellcolor{red!45} \textbf{72.00} & \cellcolor{red!20} \underline{75.72} & \cellcolor{red!45} \textbf{55.07} & \cellcolor{red!45} \textbf{49.94} & {66.00} & \cellcolor{red!45} \textbf{62.13} &  
73.57 &  \cellcolor{red!45} \textbf{69.83} &  \cellcolor{red!20} \underline{93.52} &  \cellcolor{red!45} \textbf{31.45} & {24.37} & \cellcolor{red!45} \textbf{41.68} &  \cellcolor{red!45} \textbf{52.52} \\

\Xhline{1.1pt}
    \end{tabular}
    \end{adjustbox}
    \caption{Quantitative comparison of our DCEvo  and existing
image fusion methods for downstream detection and segmentation tasks on the M$^3$FD, and FMB datasets. The best results are in \textbf{bold}, while the second best ones are \underline{underlined}. \textcolor{red!45}{Red}, \textcolor{red!20}{light red}, \textcolor{tgreen!100}{blue}, and \textcolor{tgreen!50}{light blue} represent the best, second-best, worst, and second-worst results, respectively.}
    \label{tab:odss}
\end{table*}

\begin{figure*}[!h]
  \centering 
\includegraphics[width=\linewidth, height=150pt, trim={10pt 0pt 5pt 0pt}]{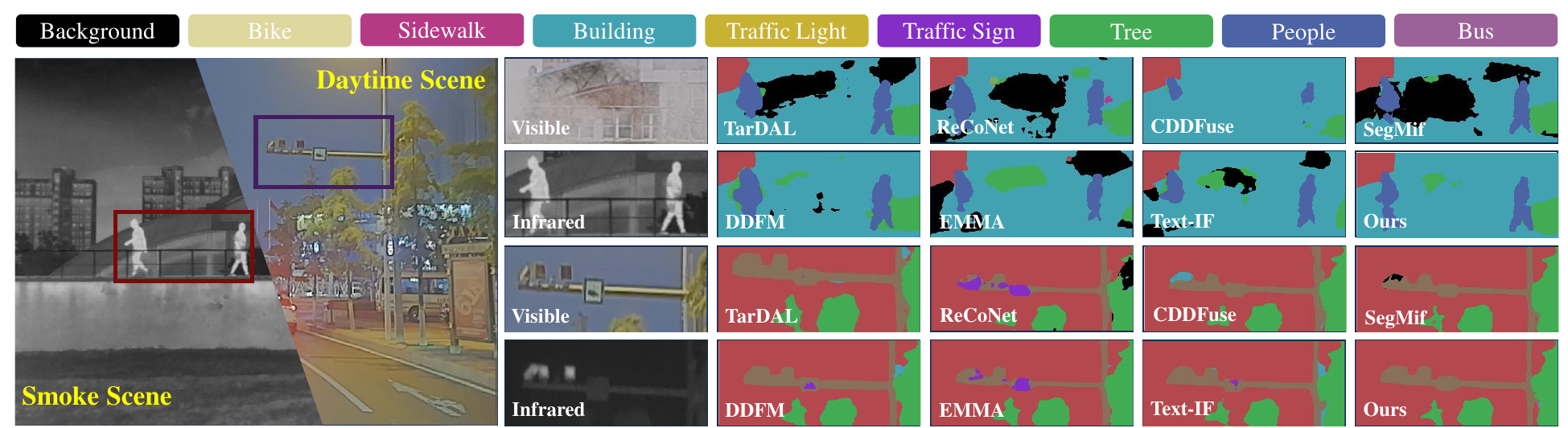}

   \caption{Qualitative comparison of our DCEvo with the fusion images generated by different fusion methods on the FMB dataset. Our approach performs the best segmentation results in the smock scene and daytime scene.
}
   \label{fig:segvis}
\end{figure*}

\begin{figure*}[t]
  \centering
\includegraphics[width=\linewidth, height=120pt,trim={10pt 5pt 5pt 5pt}]{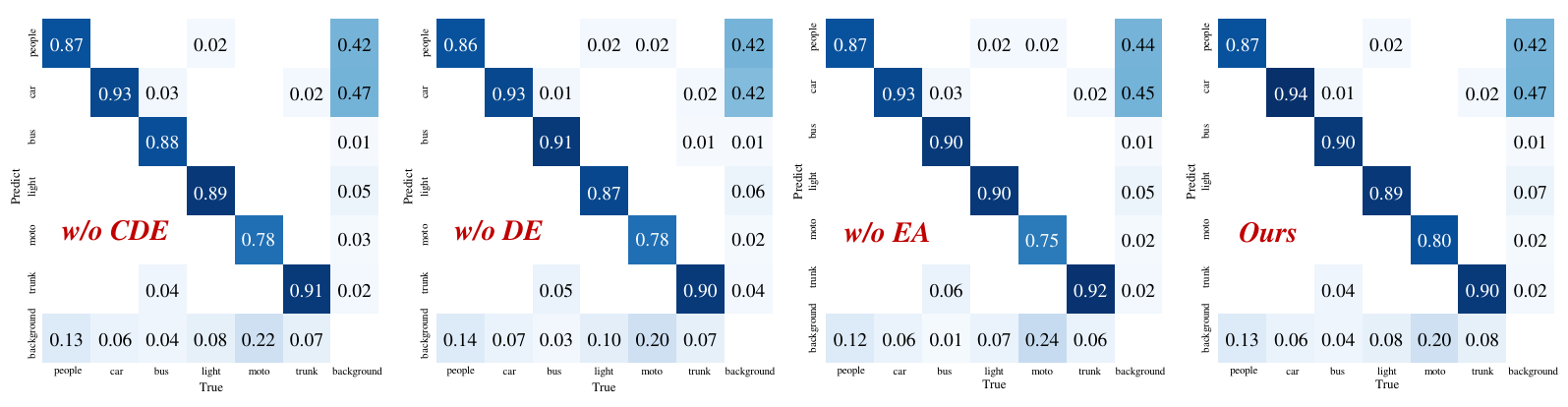}

   \caption{Normalized confusion matrix of our proposed methods in object detection. Our method has the best overall precision.
}
   \label{fig:figConfMtxAblation}
\end{figure*}



\subsection{Task-driven downstream IVIF applications}
We split the M\(^{3}\)FD dataset with an 80:20 ratio and utilize YOLOv8s for object detection. As shown in Table~\ref{tab:odss} (left), most fusion methods achieve higher mAP than using only visible or infrared images individually, with our DCEvo method attaining the highest mAP. Figure~\ref{fig:detvis} demonstrates that our fused images detect all objects in the scene, whereas other methods miss some detections. Additionally, DCEvo excels in challenging environments by leveraging DE and CDE to effectively refine and process image features.

In semantic segmentation, we use FMB dataset for training and testing.
We use Segformer-b1 as the segmentation models, for all  methods.
The results of the semantic segmentation mIoU is shown in the right of Table~\ref{tab:odss}. Notably, our DCEvo  outperforms others in terms of mIoU, achieving a 1.4–7.2\% improvement over compared approaches. As to qualitative comparisons of segmentation in Figure~\ref{fig:segvis}, our method achieves the best segmentation results.

\subsection{Ablation study}

We conduct ablation experiments to evaluate the effectiveness of the three methods. Figure~\ref{fig:figConfMtxAblation} presents the normalized confusion matrices, demonstrating that our complete model achieves the highest precision.

\begin{figure}[h]
  \centering
\includegraphics[width=\linewidth, height=105pt,trim={10pt 5pt 5pt 0pt}]{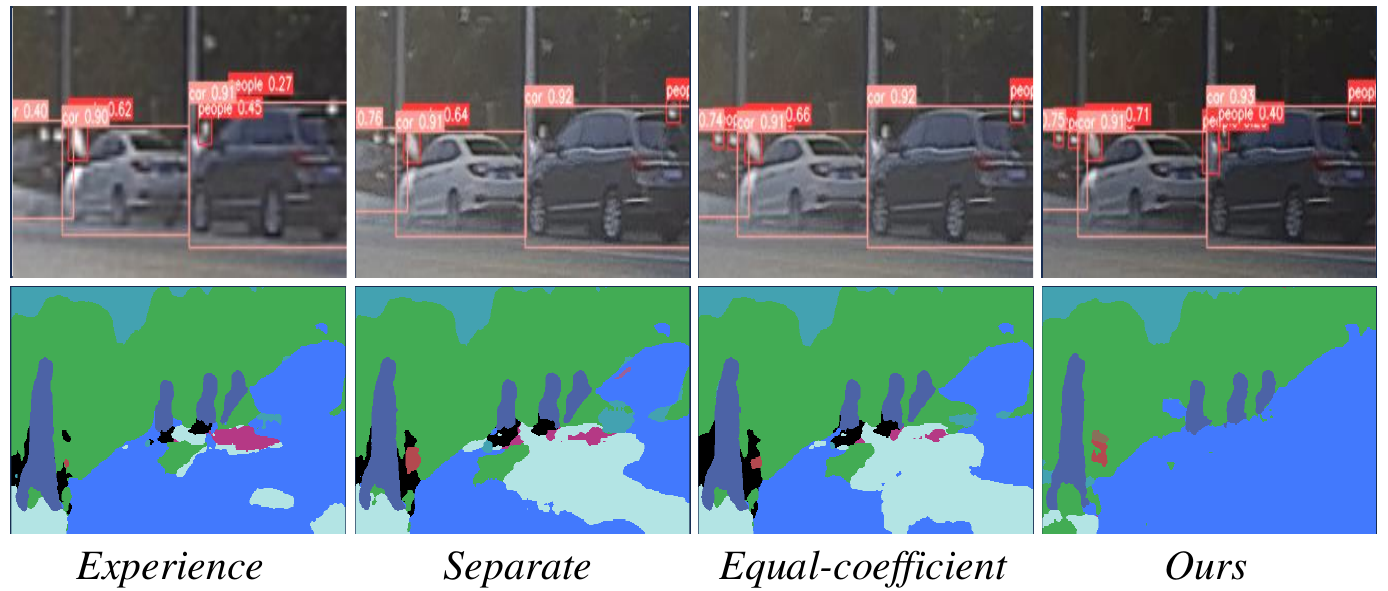}

   \caption{Qualitative  results of our evolutionary algorithm and other training methods. Our  model shows the best results.
}
   \label{fig:ablaga}
\end{figure}

\noindent\textbf{Study on evolutionary algorithm}. 
Table~\ref{tab:ablationEA} presents our ablation experiments for the proposed evolutionary learning strategy. We compare our evolutionary algorithm with hyperparameter settings based on experience and uniform values, as well as our cooperative training method for optimizing both the fusion and detection networks versus separate optimization. Our method achieves the highest quantitative metrics, and the qualitative results in Figure~\ref{fig:ablaga} demonstrate superior precision.

\begin{figure}[t]
  \centering
\includegraphics[width=\linewidth, height=135pt, trim={10pt 5pt 10pt 5pt}]{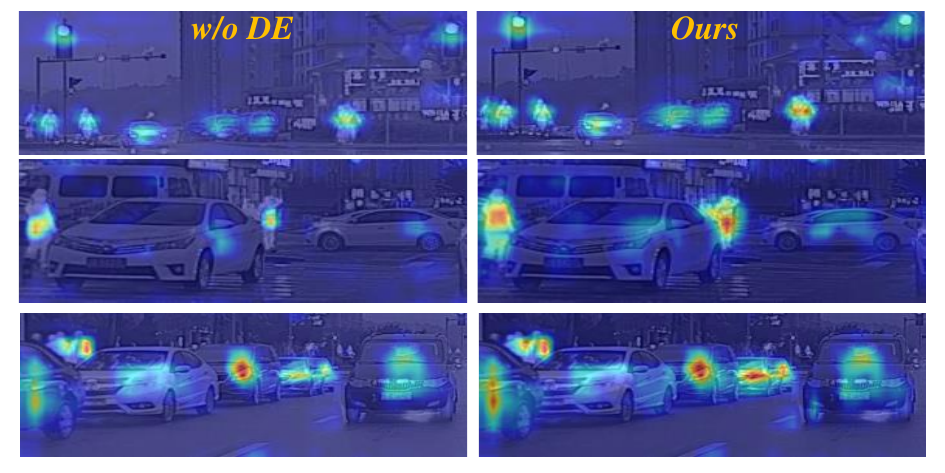}

   \caption{Comparison of the visualized feature activations of our DE.  The features
are filtered by the object detection backbone and shown by Grad-CAM. Our DE helps the network focus on object regions which are close to the ground truth.
}
   \label{fig:figPPheatmap}
\end{figure}

\noindent\textbf{Study on model architectures}. 
Our DCEvo model incorporates a discriminative enhancer and a cross-dimensional feature embedding module. To assess these enhancements, we conducted ablation experiments presented in Table~\ref{tab:ablationMS}. SP, DE, and DP denote the simple model, encoder-based discriminative enhancer, and projection-based discriminative enhancer, respectively. Base, Cat, CA, and Ours refer to the base fusion model, concatenation-based feature embedding, cross-attention only, and our cross- and self-attention feature embedding methods. DCEvo achieves the highest performance on downstream tasks, demonstrating the robustness of our architecture. Additionally, feature activations validate effectiveness of DE, as shown in Figure~\ref{fig:figPPheatmap}.

\begin{table}[t]
    \centering
\footnotesize
    \begin{adjustbox}{max width=\textwidth}
    \begin{tabular}{c|cc|cc}
\Xhline{1.1pt}
\multicolumn{1}{c|}{Metrics} 	
& \multicolumn{2}{c|}{mAP} & \multicolumn{2}{c}{mIoU} \\
        \hline
\multicolumn{1}{c|}{Methods} 
& \hspace{0.3cm}Sp\hspace{0.3cm} & \hspace{0.3cm}Co\hspace{0.3cm} & \hspace{0.3cm}Sp\hspace{0.3cm} & \hspace{0.3cm}Co\hspace{0.3cm} \\
        \hline
experience				
& \hspace{0.3cm}61.70\hspace{0.3cm} & \hspace{0.3cm}61.73\hspace{0.3cm} & \hspace{0.3cm}51.53\hspace{0.3cm} & \hspace{0.3cm}51.58\hspace{0.3cm} \\
equals					
& \hspace{0.3cm}61.53\hspace{0.3cm} & \hspace{0.3cm}61.59\hspace{0.3cm} & \hspace{0.3cm}51.27\hspace{0.3cm} & \hspace{0.3cm}51.33\hspace{0.3cm} \\
Ours						
& \hspace{0.3cm} /\hspace{0.3cm} & \hspace{0.3cm}\textbf{62.13}\hspace{0.3cm} & \hspace{0.3cm}/\hspace{0.3cm} & \textbf{52.52}\hspace{0.3cm}\\

\Xhline{1.1pt}
    \end{tabular}
\end{adjustbox}
    \caption{Quantitative comparison of our evolutionary algorithm with other training methods. \textbf{Bold} shows the best results.}
\label{tab:ablationEA}
\end{table}

\begin{table}[t]
    \centering
\footnotesize
    \begin{adjustbox}{max width=\textwidth}
    \begin{tabular}{c|ccc}
\Xhline{1.1pt}
\multicolumn{1}{c|} {Metrics} & \multicolumn{3}{c}{mAP/mIoU}\\
        \hline
\multicolumn{1}{c|}{Methods} & \hspace{0.25cm}SP\hspace{0.25cm} & \hspace{0.25cm}DE\hspace{0.25cm} & \hspace{0.25cm}DE + DP\hspace{0.25cm} \\
        \hline
Base	& \hspace{0.25cm}61.23/50.23\hspace{0.25cm} & \hspace{0.25cm}61.56/50.89\hspace{0.25cm} & 61.62/51.65\hspace{0.25cm}\\
Cat	& \hspace{0.25cm}61.29/50.55\hspace{0.25cm} & \hspace{0.25cm}61.63/51.13\hspace{0.25cm} & \hspace{0.25cm}61.67/51.78\hspace{0.25cm} \\
CA	& / & \hspace{0.25cm}61.65/51.68\hspace{0.25cm} & \hspace{0.25cm}61.70/51.92\hspace{0.25cm}  \\
Ours	& / &  \hspace{0.25cm}61.88/52.24\hspace{0.25cm} & \hspace{0.25cm}\textbf{62.13}/\textbf{52.52}\hspace{0.25cm}  \\

\Xhline{1.1pt}
    \end{tabular}
\end{adjustbox}
    \caption{Ablation results for the proposed cross-dimensional  embedding approach. \textbf{Bold} shows the best results.}
\label{tab:ablationMS}
\end{table}

\section{Concluding Remarks}
Within this paper, we introduce the Discriminative Cross-Dimensional Evolutionary Learning Framework (DCEvo) for infrared and visible image fusion, effectively enhancing both visual quality and task accuracy. By integrating discriminative enhancers and a cross-dimensional feature embedding block with an evolutionary learning strategy, DCEvo optimizes feature interactions and balances dual objectives. Experimental results across three tasks' benchmarks demonstrate that our approach surpasses state-of-the-art methods, offering robust solutions for complex environments and intelligent systems.

\textbf{Broader Impacts.}
DCEvo introduces an evolutionary learning paradigm for IVIF, simultaneously enhancing visual quality and task performance, positioning it as a new direction for perception in modern intelligent systems. Its innovative design leverages evolutionary adaptive hyperparameters to address optimization challenges across multi-dimensional tasks, improving the practicability and efficiency of integrated vision applications.

\section*{Acknowledgments}
This work is partially supported by the National Natural Science Foundation of China (Nos.62302078, 62372080, 62450072, U22B2052), the Distinguished Youth Funds of the Liaoning Natural Science Foundation (No.2025JH6/101100001), the Distinguished Young Scholars Funds of Dalian (No.2024RJ002), the China Postdoctoral Science Foundation (No.2023M730741) and the Fundamental Research Funds for the Central Universities.

{
    \small
    \bibliographystyle{ieeenat_fullname}
    \bibliography{main}

\begin{thebibliography}{49}
\providecommand{\natexlab}[1]{#1}
\providecommand{\url}[1]{\texttt{#1}}
\expandafter\ifx\csname urlstyle\endcsname\relax
  \providecommand{\doi}[1]{doi: #1}\else
  \providecommand{\doi}{doi: \begingroup \urlstyle{rm}\Url}\fi

\bibitem[Abd~Algani et~al.(2023)Abd~Algani, Caro, Bravo, Kaur, Al~Ansari, and
  Bala]{Alpher10}
Yousef~Methkal Abd~Algani, Orlando Juan~Marquez Caro, Liz Maribel~Robladillo
  Bravo, Chamandeep Kaur, Mohammed~Saleh Al~Ansari, and B~Kiran Bala.
\newblock Leaf disease identification and classification using optimized deep
  learning.
\newblock \emph{Measurement: Sensors}, 25:\penalty0 100643, 2023.

\bibitem[Arif and Wang(2020)]{Alpher15}
Muhammad Arif and Guojun Wang.
\newblock Fast curvelet transform through genetic algorithm for multimodal
  medical image fusion.
\newblock \emph{Soft Computing}, 24\penalty0 (3):\penalty0 1815--1836, 2020.

\bibitem[Coello(2007)]{Alpher16}
Carlos A~Coello Coello.
\newblock \emph{Evolutionary algorithms for solving multi-objective problems}.
\newblock Springer, 2007.

\bibitem[El~Ahmar et~al.(2022)El~Ahmar, Kolhatkar, Nowruzi, AlGhamdi, Hou, and
  Laganiere]{El_Ahmar_2022_CVPR}
Wassim~A. El~Ahmar, Dhanvin Kolhatkar, Farzan~Erlik Nowruzi, Hamzah AlGhamdi,
  Jonathan Hou, and Robert Laganiere.
\newblock Multiple object detection and tracking in the thermal spectrum.
\newblock In \emph{Proceedings of the IEEE/CVF Conference on Computer Vision
  and Pattern Recognition Workshops}, pages 277--285, 2022.

\bibitem[Geng et~al.(2024)Geng, Zhu, Wang, Zhang, Xiong, and
  Tian]{Geng_2024_CVPR}
Mengyue Geng, Lin Zhu, Lizhi Wang, Wei Zhang, Ruiqin Xiong, and Yonghong Tian.
\newblock Event-based visible and infrared fusion via multi-task collaboration.
\newblock In \emph{Proceedings of the IEEE/CVF Conference on Computer Vision
  and Pattern Recognition (CVPR)}, pages 26929--26939, 2024.

\bibitem[He et~al.(2022)He, Jiang, Lam, and Sun]{Alpher13}
Peilan He, Guiyuan Jiang, Siew-Kei Lam, and Yidan Sun.
\newblock Ml-mmas: Self-learning ant colony optimization for multi-criteria
  journey planning.
\newblock \emph{Information Sciences}, 609:\penalty0 1052--1074, 2022.

\bibitem[Holland(1992)]{Alpher06}
John~H Holland.
\newblock \emph{Adaptation in natural and artificial systems: an introductory
  analysis with applications to biology, control, and artificial intelligence}.
\newblock MIT press, 1992.

\bibitem[Huang et~al.(2022)Huang, Liu, Fan, Liu, Zhong, and
  Luo]{ReCoNet_ECCV2022}
Zhanbo Huang, Jinyuan Liu, Xin Fan, Risheng Liu, Wei Zhong, and Zhongxuan Luo.
\newblock Reconet: Recurrent correction network for fast and efficient
  multi-modality image fusion.
\newblock In \emph{Computer Vision -- ECCV 2022}, pages 539--555, Cham, 2022.
  Springer Nature Switzerland.

\bibitem[Kaur and Singh(2021)]{Alpher14}
Manjit Kaur and Dilbag Singh.
\newblock Multi-modality medical image fusion technique using multi-objective
  differential evolution based deep neural networks.
\newblock \emph{Journal of Ambient Intelligence and Humanized Computing},
  12\penalty0 (2):\penalty0 2483--2493, 2021.

\bibitem[Kavitha et~al.(2023)Kavitha, Jothi, Saravanan, Swain, Gonz{\'a}les,
  Bhardwaj, and Adomako]{Alpher11}
R Kavitha, D~Kiruba Jothi, K Saravanan, Mahendra~Pratap Swain, Jos{\'e}
  Luis~Arias Gonz{\'a}les, Rakhi~Joshi Bhardwaj, and Elijah Adomako.
\newblock [retracted] ant colony optimization-enabled cnn deep learning
  technique for accurate detection of cervical cancer.
\newblock \emph{BioMed Research International}, 2023\penalty0 (1):\penalty0
  1742891, 2023.

\bibitem[Kennedy and Eberhart(1995)]{Alpher08}
James Kennedy and Russell Eberhart.
\newblock Particle swarm optimization.
\newblock In \emph{Proceedings of ICNN'95-international conference on neural
  networks}, pages 1942--1948. ieee, 1995.

\bibitem[Kilicarslan et~al.(2021)Kilicarslan, Celik, and Sahin]{Alpher09}
Serhat Kilicarslan, Mete Celik, and {\c{S}}afak Sahin.
\newblock Hybrid models based on genetic algorithm and deep learning algorithms
  for nutritional anemia disease classification.
\newblock \emph{Biomedical Signal Processing and Control}, 63:\penalty0 102231,
  2021.

\bibitem[Li and Wu(2018)]{Alpher37}
Hui Li and Xiao-Jun Wu.
\newblock Densefuse: A fusion approach to infrared and visible images.
\newblock \emph{IEEE Transactions on Image Processing}, 28\penalty0
  (5):\penalty0 2614--2623, 2018.

\bibitem[Li et~al.(2021)Li, Wu, and Kittler]{Alpher36}
Hui Li, Xiao-Jun Wu, and Josef Kittler.
\newblock Rfn-nest: An end-to-end residual fusion network for infrared and
  visible images.
\newblock \emph{Information Fusion}, 73:\penalty0 72--86, 2021.

\bibitem[Li et~al.(2023{\natexlab{a}})Li, Xu, Wu, Lu, and
  Kittler]{LRRNet_TPRMI2023}
Hui Li, Tianyang Xu, Xiao-Jun Wu, Jiwen Lu, and Josef Kittler.
\newblock Lrrnet: A novel representation learning guided fusion network for
  infrared and visible images.
\newblock \emph{IEEE Transactions on Pattern Analysis and Machine
  Intelligence}, 45\penalty0 (9):\penalty0 11040--11052, 2023{\natexlab{a}}.

\bibitem[Li et~al.(2023{\natexlab{b}})Li, Lv, Wang, Li, Yang, and
  Yang]{DFL_2023}
Xiang Li, Chengqi Lv, Wenhai Wang, Gang Li, Lingfeng Yang, and Jian Yang.
\newblock Generalized focal loss: Towards efficient representation learning for
  dense object detection.
\newblock \emph{IEEE Transactions on Pattern Analysis and Machine
  Intelligence}, 45\penalty0 (3):\penalty0 3139--3153, 2023{\natexlab{b}}.

\bibitem[Li et~al.(2024)Li, Liu, Chen, Zou, Ma, Fan, and Liu]{li2024contourlet}
Xingyuan Li, Jinyuan Liu, Zhixin Chen, Yang Zou, Long Ma, Xin Fan, and Risheng
  Liu.
\newblock Contourlet residual for prompt learning enhanced infrared image
  super-resolution.
\newblock In \emph{European Conference on Computer Vision}, pages 270--288.
  Springer, 2024.

\bibitem[Liu et~al.(2022{\natexlab{a}})Liu, Fan, Huang, Wu, Liu, Zhong, and
  Luo]{TarDAL_CVPR2022}
Jinyuan Liu, Xin Fan, Zhanbo Huang, Guanyao Wu, Risheng Liu, Wei Zhong, and
  Zhongxuan Luo.
\newblock Target-aware dual adversarial learning and a multi-scenario
  multi-modality benchmark to fuse infrared and visible for object detection.
\newblock In \emph{Proceedings of the IEEE/CVF Conference on Computer Vision
  and Pattern Recognition}, pages 5802--5811, 2022{\natexlab{a}}.

\bibitem[Liu et~al.(2022{\natexlab{b}})Liu, Fan, Jiang, Liu, and Luo]{9349250}
Jinyuan Liu, Xin Fan, Ji Jiang, Risheng Liu, and Zhongxuan Luo.
\newblock Learning a deep multi-scale feature ensemble and an edge-attention
  guidance for image fusion.
\newblock \emph{IEEE Transactions on Circuits and Systems for Video
  Technology}, 32\penalty0 (1):\penalty0 105--119, 2022{\natexlab{b}}.

\bibitem[Liu et~al.(2023)Liu, Liu, Wu, Ma, Liu, Zhong, Luo, and
  Fan]{SegMIF_ICCV2023}
Jinyuan Liu, Zhu Liu, Guanyao Wu, Long Ma, Risheng Liu, Wei Zhong, Zhongxuan
  Luo, and Xin Fan.
\newblock Multi-interactive feature learning and a full-time multi-modality
  benchmark for image fusion and segmentation.
\newblock In \emph{Proceedings of the IEEE/CVF International Conference on
  Computer Vision}, pages 8115--8124, 2023.

\bibitem[Liu et~al.(2024{\natexlab{a}})Liu, Li, Wang, Jiang, Zhong, Fan, and
  Xu]{liu2024promptfusion}
Jinyuan Liu, Xingyuan Li, Zirui Wang, Zhiying Jiang, Wei Zhong, Wei Fan, and
  Bin Xu.
\newblock Promptfusion: Harmonized semantic prompt learning for infrared and
  visible image fusion.
\newblock \emph{IEEE/CAA Journal of Automatica Sinica}, 2024{\natexlab{a}}.

\bibitem[Liu et~al.(2024{\natexlab{b}})Liu, Lin, Wu, Liu, Luo, and
  Fan]{liu2024coconet}
Jinyuan Liu, Runjia Lin, Guanyao Wu, Risheng Liu, Zhongxuan Luo, and Xin Fan.
\newblock Coconet: Coupled contrastive learning network with multi-level
  feature ensemble for multi-modality image fusion.
\newblock \emph{International Journal of Computer Vision}, 132\penalty0
  (5):\penalty0 1748--1775, 2024{\natexlab{b}}.

\bibitem[Liu et~al.(2025)Liu, Wu, Liu, Wang, Jiang, Ma, Zhong, Fan, and
  Liu]{10812907}
Jinyuan Liu, Guanyao Wu, Zhu Liu, Di Wang, Zhiying Jiang, Long Ma, Wei Zhong,
  Xin Fan, and Risheng Liu.
\newblock Infrared and visible image fusion: From data compatibility to task
  adaption.
\newblock \emph{IEEE Transactions on Pattern Analysis and Machine
  Intelligence}, 47\penalty0 (4):\penalty0 2349--2369, 2025.

\bibitem[Liu et~al.(2021)Liu, Liu, Liu, and Fan]{Alpher38}
Risheng Liu, Zhu Liu, Jinyuan Liu, and Xin Fan.
\newblock Searching a hierarchically aggregated fusion architecture for fast
  multi-modality image fusion.
\newblock In \emph{Proceedings of the 29th ACM International Conference on
  Multimedia}, pages 1600--1608, 2021.

\bibitem[Liu et~al.(2024{\natexlab{c}})Liu, Liu, Liu, Fan, and Luo]{Alpher46}
Risheng Liu, Zhu Liu, Jinyuan Liu, Xin Fan, and Zhongxuan Luo.
\newblock A task-guided, implicitly-searched and metainitialized deep model for
  image fusion.
\newblock \emph{IEEE Transactions on Pattern Analysis and Machine
  Intelligence}, 2024{\natexlab{c}}.

\bibitem[Ma et~al.(2020)Ma, Zhang, Shao, Liang, and Xu]{Alpher39}
Jiayi Ma, Hao Zhang, Zhenfeng Shao, Pengwei Liang, and Han Xu.
\newblock Ganmcc: A generative adversarial network with multiclassification
  constraints for infrared and visible image fusion.
\newblock \emph{IEEE Transactions on Instrumentation and Measurement},
  70:\penalty0 1--14, 2020.

\bibitem[Ma et~al.(2022{\natexlab{a}})Ma, Tang, Fan, Huang, Mei, and
  Ma]{Alpher35}
Jiayi Ma, Linfeng Tang, Fan Fan, Jun Huang, Xiaoguang Mei, and Yong Ma.
\newblock Swinfusion: Cross-domain long-range learning for general image fusion
  via swin transformer.
\newblock \emph{IEEE/CAA Journal of Automatica Sinica}, 9\penalty0
  (7):\penalty0 1200--1217, 2022{\natexlab{a}}.

\bibitem[Ma et~al.(2022{\natexlab{b}})Ma, Tang, Fan, Huang, Mei, and
  Ma]{Ma_SwinFusion}
Jiayi Ma, Linfeng Tang, Fan Fan, Jun Huang, Xiaoguang Mei, and Yong Ma.
\newblock Swinfusion: Cross-domain long-range learning for general image fusion
  via swin transformer.
\newblock \emph{IEEE/CAA Journal of Automatica Sinica}, 9\penalty0
  (7):\penalty0 1200--1217, 2022{\natexlab{b}}.

\bibitem[Maniezzo et~al.(2004)Maniezzo, Gambardella, De~Luigi,
  et~al.]{Alpher07}
Vittorio Maniezzo, Luca~Maria Gambardella, Fabio De~Luigi, et~al.
\newblock Ant colony optimization.
\newblock \emph{New optimization techniques in engineering}, 141:\penalty0
  101--117, 2004.

\bibitem[Nie et~al.(2021)Nie, Ma, Cao, Ding, and Zhou]{Alpher32}
Rencan Nie, Chaozhen Ma, Jinde Cao, Hongwei Ding, and Dongming Zhou.
\newblock A total variation with joint norms for infrared and visible image
  fusion.
\newblock \emph{IEEE Transactions on Multimedia}, 24:\penalty0 1460--1472,
  2021.

\bibitem[Pan et~al.(2021)Pan, Yang, and Li]{Alpher12}
Yuwen Pan, Yuanwang Yang, and Wenzao Li.
\newblock A deep learning trained by genetic algorithm to improve the
  efficiency of path planning for data collection with multi-uav.
\newblock \emph{IEEE Access}, 9:\penalty0 7994--8005, 2021.

\bibitem[Purnomo et~al.(2024)Purnomo, Gonsalves, Mailoa, Santoso, Pribadi,
  et~al.]{Alpher21}
Hindriyanto Purnomo, Tad Gonsalves, Evangs Mailoa, Fian~Julio Santoso,
  Muhammad~Rizky Pribadi, et~al.
\newblock Metaheuristics approach for hyperparameter tuning of convolutional
  neural network.
\newblock \emph{Jurnal RESTI (Rekayasa Sistem dan Teknologi Informasi)},
  8\penalty0 (3):\penalty0 340--345, 2024.

\bibitem[Raji et~al.(2022)Raji, Bello-Salau, Umoh, Onumanyi, Adegboye, and
  Salawudeen]{Alpher20}
Ismail~Damilola Raji, Habeeb Bello-Salau, Ime~Jarlath Umoh, Adeiza~James
  Onumanyi, Mutiu~Adesina Adegboye, and Ahmed~Tijani Salawudeen.
\newblock Simple deterministic selection-based genetic algorithm for
  hyperparameter tuning of machine learning models.
\newblock \emph{Applied Sciences}, 12\penalty0 (3):\penalty0 1186, 2022.

\bibitem[Tang et~al.(2023)Tang, He, and Liu]{YDTR_2023}
Wei Tang, Fazhi He, and Yu Liu.
\newblock Ydtr: Infrared and visible image fusion via y-shape dynamic
  transformer.
\newblock \emph{IEEE Transactions on Multimedia}, 25:\penalty0 5413--5428,
  2023.

\bibitem[Tian et~al.(2022)Tian, Zhang, Yu, and Ma]{Alpher34}
Xin Tian, Wei Zhang, Dian Yu, and Jiayi Ma.
\newblock Sparse tensor prior for hyperspectral, multispectral, and
  panchromatic image fusion.
\newblock \emph{IEEE/CAA Journal of Automatica Sinica}, 10\penalty0
  (1):\penalty0 284--286, 2022.

\bibitem[Toet and Hogervorst(2012)]{TNODATA}
Alexander Toet and Maarten~A. Hogervorst.
\newblock {Progress in color night vision}.
\newblock \emph{Optical Engineering}, 51\penalty0 (1):\penalty0 010901, 2012.

\bibitem[Wang et~al.(2023)Wang, Li, Chen, Zhang, and Zhan]{Alpher19}
Ye-Qun Wang, Jian-Yu Li, Chun-Hua Chen, Jun Zhang, and Zhi-Hui Zhan.
\newblock Scale adaptive fitness evaluation-based particle swarm optimisation
  for hyperparameter and architecture optimisation in neural networks and deep
  learning.
\newblock \emph{CAAI Transactions on Intelligence Technology}, 8\penalty0
  (3):\penalty0 849--862, 2023.

\bibitem[Xu et~al.(2020{\natexlab{a}})Xu, Ma, Le, Jiang, and
  Guo]{RoadSceneDATA}
Han Xu, Jiayi Ma, Zhuliang Le, Junjun Jiang, and Xiaojie Guo.
\newblock Fusiondn: A unified densely connected network for image fusion.
\newblock \emph{Proceedings of the AAAI Conference on Artificial Intelligence},
  34\penalty0 (07):\penalty0 12484--12491, 2020{\natexlab{a}}.

\bibitem[Xu et~al.(2020{\natexlab{b}})Xu, Ma, and Zhang]{Alpher40}
Han Xu, Jiayi Ma, and Xiao-Ping Zhang.
\newblock Mef-gan: Multi-exposure image fusion via generative adversarial
  networks.
\newblock \emph{IEEE Transactions on Image Processing}, 29:\penalty0
  7203--7216, 2020{\natexlab{b}}.

\bibitem[Xu et~al.(2022)Xu, Ma, Jiang, Guo, and Ling]{U2_2022}
Han Xu, Jiayi Ma, Junjun Jiang, Xiaojie Guo, and Haibin Ling.
\newblock U2fusion: A unified unsupervised image fusion network.
\newblock \emph{IEEE Transactions on Pattern Analysis and Machine
  Intelligence}, 44\penalty0 (1):\penalty0 502--518, 2022.

\bibitem[Yi et~al.(2024)Yi, Xu, Zhang, Tang, and Ma]{TextIF_CVPR2024}
Xunpeng Yi, Han Xu, Hao Zhang, Linfeng Tang, and Jiayi Ma.
\newblock Text-if: Leveraging semantic text guidance for degradation-aware and
  interactive image fusion.
\newblock In \emph{Proceedings of the IEEE/CVF Conference on Computer Vision
  and Pattern Recognition}, pages 27026--27035, 2024.

\bibitem[Yu et~al.(2023)Yu, Li, Shen, Liu, and Wang]{Yu_2023_CVPR}
Zhenjie Yu, Shuang Li, Yirui Shen, Chi~Harold Liu, and Shuigen Wang.
\newblock On the difficulty of unpaired infrared-to-visible video translation:
  Fine-grained content-rich patches transfer.
\newblock In \emph{Proceedings of the IEEE/CVF Conference on Computer Vision
  and Pattern Recognition}, pages 1631--1640, 2023.

\bibitem[Zhao et~al.(2023{\natexlab{a}})Zhao, Xie, Zhao, He, and
  Lu]{MetaFusion_CVPR2023}
Wenda Zhao, Shigeng Xie, Fan Zhao, You He, and Huchuan Lu.
\newblock Metafusion: Infrared and visible image fusion via meta-feature
  embedding from object detection.
\newblock In \emph{Proceedings of the IEEE/CVF Conference on Computer Vision
  and Pattern Recognition}, pages 13955--13965, 2023{\natexlab{a}}.

\bibitem[Zhao et~al.(2021)Zhao, Xu, Zhang, Liang, Zhang, and Liu]{Alpher31}
Zixiang Zhao, Shuang Xu, Jiangshe Zhang, Chengyang Liang, Chunxia Zhang, and
  Junmin Liu.
\newblock Efficient and model-based infrared and visible image fusion via
  algorithm unrolling.
\newblock \emph{IEEE Transactions on Circuits and Systems for Video
  Technology}, 32\penalty0 (3):\penalty0 1186--1196, 2021.

\bibitem[Zhao et~al.(2023{\natexlab{b}})Zhao, Bai, Zhang, Zhang, Xu, Lin,
  Timofte, and Van~Gool]{CDDFuse_CVPR2023}
Zixiang Zhao, Haowen Bai, Jiangshe Zhang, Yulun Zhang, Shuang Xu, Zudi Lin,
  Radu Timofte, and Luc Van~Gool.
\newblock Cddfuse: Correlation-driven dual-branch feature decomposition for
  multi-modality image fusion.
\newblock In \emph{Proceedings of the IEEE/CVF Conference on Computer Vision
  and Pattern Recognition}, pages 5906--5916, 2023{\natexlab{b}}.

\bibitem[Zhao et~al.(2023{\natexlab{c}})Zhao, Bai, Zhu, Zhang, Xu, Zhang,
  Zhang, Meng, Timofte, and Van~Gool]{DDFM_2023_ICCV}
Zixiang Zhao, Haowen Bai, Yuanzhi Zhu, Jiangshe Zhang, Shuang Xu, Yulun Zhang,
  Kai Zhang, Deyu Meng, Radu Timofte, and Luc Van~Gool.
\newblock Ddfm: Denoising diffusion model for multi-modality image fusion.
\newblock In \emph{Proceedings of the IEEE/CVF International Conference on
  Computer Vision}, pages 8082--8093, 2023{\natexlab{c}}.

\bibitem[Zhao et~al.(2024)Zhao, Bai, Zhang, Zhang, Zhang, Xu, Chen, Timofte,
  and Van~Gool]{EMMA_CVPR2024}
Zixiang Zhao, Haowen Bai, Jiangshe Zhang, Yulun Zhang, Kai Zhang, Shuang Xu,
  Dongdong Chen, Radu Timofte, and Luc Van~Gool.
\newblock Equivariant multi-modality image fusion.
\newblock In \emph{Proceedings of the IEEE/CVF Conference on Computer Vision
  and Pattern Recognition}, pages 25912--25921, 2024.

\bibitem[Zheng et~al.(2024)Zheng, Zhou, Huang, Hou, Li, Xu, and
  Zhao]{Zheng_2024_CVPR}
Naishan Zheng, Man Zhou, Jie Huang, Junming Hou, Haoying Li, Yuan Xu, and Feng
  Zhao.
\newblock Probing synergistic high-order interaction in infrared and visible
  image fusion.
\newblock In \emph{Proceedings of the IEEE/CVF Conference on Computer Vision
  and Pattern Recognition}, pages 26384--26395, 2024.

\bibitem[Zhou et~al.(2011)Zhou, Qu, Li, Zhao, Suganthan, and Zhang]{Alpher17}
Aimin Zhou, Bo-Yang Qu, Hui Li, Shi-Zheng Zhao, Ponnuthurai~Nagaratnam
  Suganthan, and Qingfu Zhang.
\newblock Multiobjective evolutionary algorithms: A survey of the state of the
  art.
\newblock \emph{Swarm and evolutionary computation}, 1\penalty0 (1):\penalty0
  32--49, 2011.

\end{thebibliography}
}


\end{document}